\documentclass[review]{elsarticle}

\usepackage{bm}
\usepackage{mathrsfs}
\usepackage{subfigure}
\usepackage{booktabs}
\usepackage{amssymb}
\usepackage{amsmath}
\usepackage{graphicx}
\usepackage{epstopdf}
\usepackage{color}
\usepackage{multirow}
\usepackage{geometry}
\journal{}

\begin{document}
    \pagestyle{empty}
    \begin{frontmatter}

\title{Active noise control techniques for nonlinear systems}

\tnotetext[mytitlenote] {The work is supported by the National Science Foundation of P.R. China under Grant 61901285, 61901400, and 61701327, Sichuan Science and Technology Fund under Grant 20YYJC3709, China Postdoctoral Science Foundation under Grant 2020T130453, and Sichuan University Postdoctoral Interdisciplinary Fund.\\E-mail\;addresses: \texttt{lulu19900303@126.com(L.\;Lu), kl\_yin@hotmail.com(K.-L.\;Yin), delamare@cetuc.puc-rio.br(R.C.\;Lamare), zongsheng56@126.com(Z.\;Zheng),  yuyi\_xyuan@163.com(Y.\;Yu), arielyang@scu.edu.cn(X. Yang), chenbd@mail.xjtu.edu.cn(B. Chen)}. \\Corresponding author: Xiaomin Yang.}

\author{Lu Lu$^{a}$, Kai-Li Yin$^{b}$, Rodrigo C. de Lamare$^{c}$, Zongsheng Zheng$^{d}$, Yi Yu$^{e}$, Xiaomin Yang$^{a*}$, Badong Chen$^{f}$}
\address{a) School of Electronics and Information Engineering, Sichuan University, Chengdu, Sichuan 610065, China.}
\address{b) School of Computer Science, Sichuan University, Chengdu, Sichuan 610065, China.}
\address{c) CETUC, PUC-Rio, Rio de Janeiro 22451-900, Brazil.}
\address{d) School of Electrical Engineering, Sichuan University, Chengdu, 610065, China.}
\address{e) School of Information Engineering, Robot Technology Used for Special Environment Key Laboratory of Sichuan Province, Southwest University of Science    and Technology, Mianyang 621010, China.}
\address{f) School of Electronic and Information Engineering, Xi'an Jiaotong University, Xi'an 710049, China.}

\begin{abstract}
Most of the literature focuses on the development of the linear
active noise control (ANC) techniques. However, ANC systems might
have to deal with some nonlinear components and the performance of
linear ANC techniques may degrade in this scenario. To overcome this
limitation, nonlinear ANC (NLANC) algorithms were developed. In Part
II, we review the development of NLANC algorithms during the last
decade. The contributions of heuristic ANC algorithms are outlined.
Moreover, we emphasize recent advances of NLANC algorithms, such as
spline ANC algorithms, kernel adaptive filters, and nonlinear
distributed ANC algorithms. Then, we present recent applications of
ANC technique including linear and nonlinear perspectives. Future
research challenges regarding ANC techniques are also discussed.
\end{abstract}


\end{frontmatter}


\section{Introduction}
\label{sec:Introduction}

The celebrated filtered-x least mean square (FxLMS) algorithm and
its variants have been successfully applied in active noise control
(ANC) systems. However, in practical situations, the primary path
$P(z)$ and secondary path $S(z)$ may be nonlinear
\cite{das2004active}. Moreover, the reference noise $d(n)$ arises
from dynamic systems and as such, the noise may be a nonlinear and
deterministic or stochastic, colored, and non-Gaussian signal
\cite{george2013review}. In these cases, linear ANC techniques
cannot fully utilize the coherence in the noise and achieve
suboptimal performance.

\begin{table}[h]
    \scriptsize
    \centering \doublerulesep=0.05pt
    \caption{Time line of NLANC.}
    \begin{tabular}{l|l|l|l}
        \hline\hline
        \textbf{Years} & \textbf{Authors} & \textbf{Contributions} & \textbf{References} \\ \hline
        1994      &Wangler and Hansen         &\begin{tabular}[c]{@{}l@{}}Genetic algorithm for nonlinear\\active noise and vibration control\end{tabular}               & \cite{wangler1994genetic}            \\ \hline
        1995      &Snyder and Tanaka         &ANN for NLANC               &\begin{tabular}[c]{@{}l@{}} \cite{snyder1995active}\end{tabular}            \\ \hline
        1996      &Tang et al.          &\begin{tabular}[c]{@{}l@{}}Genetic algorithm for ANC\end{tabular}                &\cite{tang1996application}            \\ \hline
        1997      &Tokhi and Wood          &\begin{tabular}[c]{@{}l@{}}Radial basis function networks for NLANC\end{tabular}                &\cite{tokhi1997active}            \\ \hline
        1998      &Strauch and Mulgrew          &\begin{tabular}[c]{@{}l@{}}A combined linear and
            nonlinear controller to\\compensate for the nonlinear effects\end{tabular}                & \cite{strauch1998active}            \\ \hline
        1999      &\begin{tabular}[c]{@{}l@{}}Bouchard, Paillard,\\and Dinh\end{tabular}          &\begin{tabular}[c]{@{}l@{}}Five new algorithms for the learning of a\\multilayer perceptron neural network for NLANC\end{tabular}                &\cite{bouchard1999improved}            \\ \hline
        1997, 2001    &Tan and Jiang         &VFxLMS algorithm        & \cite{tan1997filtered,tan2001adaptive}            \\ \hline
        2001    &Bouchard         &\begin{tabular}[c]{@{}l@{}}Filtered-neuron-level extended Kalman\\algorithm (FxNEKA) and filtered-x extended\\Kalman filter (FxEKF) for NLANC\end{tabular}        &\cite{bouchard2001new}            \\ \hline
        2004      &Das and Panda         &FsLMS algorithm               &\cite{das2004active}            \\ \hline
        2004      &Sicuranza and Carini         &\begin{tabular}[c]{@{}l@{}}Volterra filtered-x affine projection\\ (VFxAP) algorithm\end{tabular}                & \cite{sicuranza2004filtered}            \\ \hline
        2005      &Sicuranza and Carini         &\begin{tabular}[c]{@{}l@{}}Implementation of  the VFxAP algorithm\\based on V-vector algebra\end{tabular}     & \cite{sicuranza2005multichannel}            \\ \hline
        2005      &Kuo and Wu         &\begin{tabular}[c]{@{}l@{}}Bilinear FxLMS algorithm\end{tabular}                &\cite{kuo2005nonlinear}           \\ \hline
        2006      &\begin{tabular}[c]{@{}l@{}}Modares, Ahmadyfard,\\ and Hadadzarif \end{tabular}          &\begin{tabular}[c]{@{}l@{}}PSO for NLANC\end{tabular}                & \cite{modares2006pso}            \\ \hline
        2004, 2006      &Zhang et al.          &\begin{tabular}[c]{@{}l@{}}Fuzzy neural networks for NLANC\end{tabular}     & \cite{zhang2004active,zhang2006adaptive}            \\ \hline
        2006      &Sicuranza and Carini         &\begin{tabular}[c]{@{}l@{}}Generalized Hammerstein model for NLANC\end{tabular}     & \cite{sicuranza2006accuracy}            \\ \hline
        2007      &Chang and Luoh          &\begin{tabular}[c]{@{}l@{}}Neural-based FxLMS algorithm\end{tabular}                &\cite{chang2007enhancement}            \\ \hline
        2008      &Bambang         &\begin{tabular}[c]{@{}l@{}}Recurrent neural networks for NLANC\end{tabular}     & \cite{bambang2008adjoint}            \\ \hline
        2008      &\begin{tabular}[c]{@{}l@{}}Reddy, Das, \\and
            Prabhu\end{tabular}          &Fast FsLMS algorithm             &\cite{reddy2008fast} \\ \hline\hline
    \end{tabular}
    \label{Table000}
\end{table}

To address these limitations, several nonlinear ANC (NLANC)
algorithms were proposed, yielding a more stable performance and
lower noise level
\cite{wangler1994genetic,sicuranza2004filtered,reddy2008fast}.  It
should be emphasized that some NLANC methods mentioned in this paper
first appeared before 2009. The research on NLANC dates back to
1995, 59 years after the concept of ANC was proposed by Lueg
\cite{paul1936process}. In this year, Snyder and Tanaka developed
ANN for NLANC problem, and so far a large number of NLANC algorithms
have been proposed. The milestones in the progress of NLANC before
2009 are presented in Table \ref{Table000}. Survey papers on NLANC
techniques have been reported \cite{george2013review}. However,
these surveys do not cover the literature since 2013. To complete
the review of NLANC techniques and include the advances, in this
paper we review the research progress of NLANC models in recent
years.

The Volterra filter is gaining importance in NLANC. According to the
\textit{Stone-Weierstrass approximation theorem}, the Volterra
filter is a `universal approximator', i.e., the filter can
arbitrarily approximate discrete-time with fading memory,
time-invariant, continuous, finite-memory, nonlinear system
\cite{burt2018efficient}. The Volterra filter belongs to
linear-in-the-parameters (LIP) filters, where the output depends
linearly on the parameters of the filter itself. Note that this type
of filter becomes computationally expensive when a large number of
coefficients are required \cite{burt2018efficient}. To solve this
problem, the second-order Volterra (SOV) series and the third-order
Volterra (TOV) series are often used. For example, in
\cite{lu2016adaptive,he2019an}, the Volterra filtered-x-based
algorithms that use of SOV were developed for active impulsive noise
control (AINC) problem.

The NLANC based on an artificial neural network (ANN) method was
first studied, which provides the capacity to maintain causality
within the control scheme. From 1997 to 2008, many papers discussed
the usage of radial basis function (RBF) networks, recurrent neural
networks (RNNs), and fuzzy neural networks for NLANC (see Table
\ref{Table000}). In the last decade, some attempts were made by
using an ANN as the controller
\cite{le2017adaptive,zhang2010adaptive,azadi2012filtered}. A neural
controller equipped with the filtered-u least mean square (FuLMS)
algorithm with correction terms momentum was proposed in
\cite{chang2010neural}. The correction terms momentum can
efficiently prevent the system from unstable poles and can enhance
the performance as compared with the conventional leaky FuLMS
algorithm.  Regrettably, the above mentioned ANNs have heavy
computational complexity which may hinder their practical
applications \cite{george2013review}.

The functional link ANN (FLANN) exploits the single layer of ANN to
obtain reliable estimation performance with acceptable error level
\cite{reddy2009fast}. The FLANN filter is also a type of LIP filter,
which shares a unified implementation with the Volterra filter.
Moreover, the linear adaptive filtering algorithms can easily extend
to this framework with favorable implications on the algorithm
stability and computational complexity \cite{sicuranza2011a}. Over
the past decade, a number of filtered-s least mean square (FsLMS)
algorithms were proposed by employing FLANN, such as fast FsLMS
\cite{reddy2009fast}, robust FsLMS (RFsLMS) \cite{george2012a}. Most
of these algorithms use trigonometric functional expansion as the
basis function, and a small number of algorithms that consider using
Chebyshev \cite{chen2019nonlinear}, Fourier \cite{carini2012new},
and Legendre expansions \cite{george2013active} were also developed.

To compensate for the strong nonlinearity of  the system with
saturated signals, several bilinear algorithms with reduced
computational complexity were proposed, which employ the cross-terms
based on both feedback and feedforward polynomials and they can
model NLANC systems with shorter filter length
\cite{le2019m-max,guo2018adaptive}. In \cite{le2018bilinear}, a new
bilinear filter was proposed, which uses FLANN to expand the input
vector and then adapts coefficients via bilinear filter to obtain
excellent nonlinear modeling capability.

In this context, the particle swarm optimization (PSO) algorithm is
an effective biologically inspired meta-heuristic algorithm, which
performs by a competition and cooperation mechanism between the
particle swarm and thus it retains the global search strategy of
population \cite{rout2016particle}. Such algorithm has been
extensively studied in ANC technique, especially for nonlinear
system \cite{rout2012particle}. The most obvious advantage of the
PSO-ANC algorithm is that it can reduce noise without online
estimation of the secondary path
\cite{rout2012particle,rout2019pso}. In addition, other heuristic
ANC algorithms were also developed, see \cite{khan2019a} and
references therein.

Since 2009, several attractive methods were proposed for NLANC. For
performance improvement, the kernel adaptive filter (KAF) algorithm
and spline adaptive filter were introduced
\cite{bao2009active,scarpiniti2013nonlinear}. The KAF recasts the
input data into a high-dimensional feature space via a reproducing
kernel Hilbert space (RKHS) and as such, the linear adaptive filter
is applied in the feature space \cite{bao2009active}. In contrast,
the spline adaptive filter establishes a nonlinear mathematics model
by an adaptive look-up table (LUT) in which the control points are
interpolated by a local low-order polynomial spline curve
\cite{scarpiniti2013nonlinear}. An important merit of the spline
adaptive filter is that its computational complexity is much lower
as compared to other traditional NLANC algorithms
\cite{patel2016compensating}. Motivated by distributed algorithms,
some nonlinear distributed algorithms were proposed by integrating
FLANN into distributed ANC algorithms, which is important for
ensuring that the wireless acoustic sensor networks (WASNs) can
combat nonlinear distortions \cite{kukde2017on}.

Initial studies have shown that the KAF is limited by high
computational complexity. Specifically, the input data of KAF must
be a dictionary, where every new data input that arrives is used to
calculate the filter output \cite{Weifeng2010Kernel}. To overcome
this limitation, the quantized scheme \cite{chen2011quantized} and
the set-membership scheme \cite{flores2019set-membership} can be
taken into account to start new research on the field of NLANC. For
example, the adaptive step size scheme and data-selective update
strategy proposed in \cite{flores2019set-membership} can be directly
combined with the KAF in NLANC. On the other front, Internet of
Things (IoT) has been shown to be compatible with communication and
signal processing techniques for creating a smart world. In
\cite{shen2018mute}, the concept of ANC was introduced to IoT,
generating \textit{MUTE leverages}, for environmental noise
cancellation. In \cite{galambos2015active}, the concept system of
ANC-IoT is tested. All these works have demonstrated the feasibility
and great potential of the combination of the IoT and ANC.

The goal of this survey is to provide a comprehensive overview of
the latest NLANC research results and innovations. We also discuss
the applications of ANC techniques that have been introduced in many
fields, such as functional magnetic resonance imaging (fMRI)
acoustic noise control \cite{reddy2011hybrid} and transformer noise
control \cite{zhang2012active}. Moreover, we present some remarks on
future research directions for ANC. The papers that have been
published since 2009 are considered in this survey. In particular,
the main problems that the NLANC algorithms should address are
presented in Section \ref{sec:b2}. The modeling method of NLANC is
discussed in Section \ref{sec:2}, where Volterra ANC algorithms,
FLANN-based algorithms, and bilinear ANC algorithms are used as
examples. Section \ref{sec:3} focuses on
evolutionary-computing-based ANC algorithms. In Section \ref{sec:4},
some novel NLANC methods emerging in the past decade are
investigated. Some important implementation issues and future
research challenges of ANC are discussed in Sections \ref{sec:5} and
\ref{sec:6}, respectively. Finally, conclusions are provided in
Section \ref{sec:7}. All the brackets follow the order $\{ [
(\cdot)]\}$.

\section{Problem formulation}
\label{sec:b2}

In practical situations, three types of nonlinearities may exist in
ANC systems. First of all, the primary noise at the cancellation
point $d(n)$ may exhibit nonlinear effects. For instance, when the
primary noise propagates in a duct with very high sound pressure,
the transfer function of $P(z)$ can be well modeled by nonlinear
function \cite{chen2019nonlinear,luo2016efficient}.

On the other hand, the secondary plant $S(z)$ models the signal
converters (A/D and D/A), power amplifiers, and transducers
(actuators or speakers). The nonlinear effect may be caused by
overdriving the electronics or the speakers/transducers in the
secondary path \cite{napoli2009nonlinear}, and the amplifier may be
in saturation \cite{kuo2010active}. In these cases, the nonlinear
characteristics should be considered in the mathematical model for
enhanced performance. In research experiments, the transfer function
between the speaker and the error microphone is often modeled by
\textit{minimum} or \textit{non-minimum phase} response.

Apart from the above two cases, some nonlinear distortions may come
from the components of the control system. The most common source of
nonlinearity in ANC systems is the actuator. The control actuator
has typically nonlinear response characteristics, when it excites
the frequency of interest and its related harmonics at the same
time, and it is hoped that the ANC system will compensate and
overcome such adverse effect. Furthermore, the noise arises from
dynamic systems and as such, the noise may be a nonlinear and
deterministic signal \cite{lu2016adaptive}. For example, it is known
that the noise from blowers, grinders, airfoils, and fans is shown
to be chaotic \cite{das2012nonlinear,behera2014nonlinear}. The type
and examples of nonlinearities present in ANC systems are summarized
in Table \ref{Table00}.

\begin{table}[]
    \scriptsize
    \centering \doublerulesep=0.05pt
    \caption{The type and examples of nonlinearities present in ANC systems.}
    \begin{tabular}{l|l}
        \hline
        \textbf{Type of nonlinearities}                                                                & \textbf{Examples} \\ \hline
        Nonlinearities in the primary path                                                            &  \begin{tabular}[c]{@{}l@{}}The primary noise propagates in a duct with very high \\sound pressure. \end{tabular}        \\ \hline
        Nonlinearities in the secondary path                                                            &  \begin{tabular}[c]{@{}l@{}} (1)  Overdriving the electronics or the speakers/transducers \\in the secondary path.\\ (2) The amplifier may be in saturation.   \end{tabular}  \\ \hline
        \begin{tabular}[c]{@{}l@{}}Nonlinearities in the components\\ of the ANC systems\end{tabular}
        & \begin{tabular}[c]{@{}l@{}} (1) Actuator excites the frequency of interest and its \\related harmonics at the same time.\\ (2) Actuator operates with the input signal whose \\amplitude is close to or higher than the nominal input\\ signal value.\\ (3) The working frequency of the actuator is beyond \\the normal frequency range (or close to the limit).\\ (4) Noise from the blowers, grinders, airfoils, and fans\\ is shown to be chaotic. \end{tabular}       \\ \hline
    \end{tabular}
\label{Table00}
\end{table}

In these cases, the linear ANC techniques cannot fully utilize the
coherence in the noise and achieve suboptimal performance even
though most adaptive algorithms
\cite{jidf,spa,intadap,mbdf,jio,jiols,jiomimo,sjidf,ccmmwf,tds,mfdf,l1stap,mberdf,jio_lcmv,locsme,smtvb,ccmrls,dce,itic,jiostap,aifir,ccmmimo,vsscmv,bfidd,mbsic,wlmwf,bbprec,okspme,rdrcb,smce,armo,wljio,saap,vfap,saalt,mcg,sintprec,stmfdf,1bitidd,jpais,did,rrmber,memd,jiodf,baplnc,als,vssccm,doaalrd,jidfecho,dcg,rccm,ccmavf,mberrr,damdc,smjio,saabf,arh,lsomp,jrpaalt,smccm,vssccm2,vffccm,sor,aaidd,lrcc,kaesprit,lcdcd,smbeam,ccmjio,wlccm,dlmme,listmtc,smcg,mfsic,cqabd,rmmse,rsthp,dmsmtc,dynovs,dqalms,detmtc,1bitce,mwc,dlmm,rsbd,rdcoprime,rdlms,lbal,wlbd,rrser},
can be adapted to NLANC scenarios. To tackle these limitations, both
the models and the algorithms utilized in ANC must be properly
re-designed, resulting in NLANC. There are two problems which should
be addressed in NLANC: what are the practical performance
characteristics of nonlinear model in this implementation, and what
are the details of an algorithm which can be used to adapt a
nonlinear architecture for NLANC implementation.

\section{NLANC algorithms}
\label{sec:2}

The modeling accuracy of NLANC systems is crucial for meeting the
performance requirements of noise reduction level. This section
presents different approaches for the design of NLANC.

\begin{figure}[!htb]
    \centering
    \includegraphics[scale=0.8]{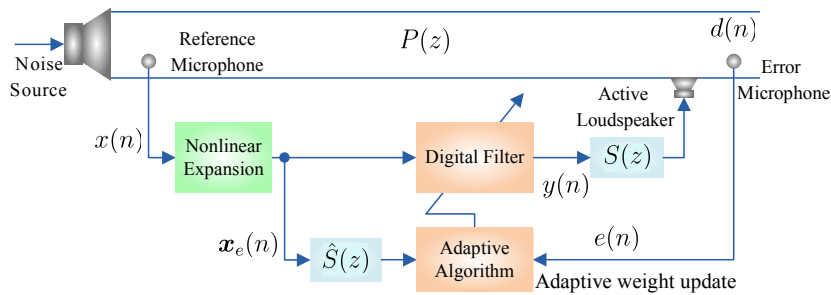}
    \caption{\label{01} Diagram of feedforward NLANC system.}
    \label{Fig01}
\end{figure}

\subsection{Volterra ANC algorithms}

Nonlinear autoregressive with exogenous variables (NARX) models have been preliminarily considered for NLANC techniques. Such methods are flexible and satisfy the representation capability of nonlinear systems. In addition, the truncated Volterra filter and bilinear filter can be subsumed into the NARX model. The potential superiority of the narrowband ANC (NANC) model for NLANC applications has been emphasized in \cite{napoli2009nonlinear} and the improved version that considers the commutation of the control filter block and the secondary path dynamics was proposed in \cite{delvecchio2014dual}.

The Volterra ANC algorithms have been studied by several researchers. The preeminent merit of this modeling approaches is that they are `universal approximators' that models diverse systems with satisfied performance. The block diagram of the NLANC system is shown in Fig. \ref{Fig01}, where $e(n)$ denotes the residual noise, $\hat S(z)$ is the estimate of $S(z)$, $x(n)$ stands for the reference signal, $d(n)$ denotes the undesired signal, and $\bm x_e(n)$ denotes the expanded input vector. The Volterra ANC system utilizes truncated Volterra series as a nonlinear expansion. The filter output $y(n)$ of these algorithms can be represented by
\begin{equation}
y(n) = \sum\limits_{q = 1}^Q {\sum\limits_{m_1 = 0}^{M-1} {\cdots\sum\limits_{m_q = 0}^{M-1} {{h_q}(m_1,...,m_q)\prod\limits_{i = 1}^q {x(n - m_i)}}}}
\label{002}
\end{equation}
where $Q$ denotes the order of nonlinearity, $M$ is the memory length, and $h_q(\cdot)$ is the $q$th-order Volterra kernel. The filter output of (\ref{002}) can be compactly rewritten as
\begin{equation}
y(n) = \bm h^{\mathrm T}\left[\bm x^{\mathrm T}_1(n),\bm x^{\mathrm T}_2(n),\ldots,\bm x^{\mathrm T}_Q(n)\right]
\label{002b}
\end{equation}
where $\bm h$ denotes the filter coefficient vector, $(\cdot)^{\mathrm T}$ is the transposition, $\bm x_1(n)=[x(n),x(n-1),\ldots,x(n-M+1)]^{\mathrm T}$, and the th$j$-order input vector $\bm x_j(n)$ can be gained recursively by
\begin{equation}
\bm x_j(n) = \bm x_1(n) \otimes \bm x_{j-1}(n),\;\;j=2,3,\ldots,Q
\label{002c}
\end{equation}
where $\otimes$ represents the Kronecker product.

In the general case with memory size $M$ and nonlinearity order $Q$, the number of coefficients  can be computed by $N_c = \frac{(M+Q)!}{M!Q!}-1$. We can observe that the number of model coefficients grows in $M^Q$, causing an exponential dimension increase. Therefore, using a truncation strategy to limit the value of $Q$ is indispensable in practice. For $Q=2$, the SOV-ANC algorithm is obtained, whose filter output can be expressed as follows \cite{lu2016adaptive}:
\begin{equation}
\begin{aligned}
y(n) = {\bm h}^\mathrm{T}{\bm x}_e(n)  = \sum\limits_{m_1 = 0}^{M-1} h_{1}(m_1)x(n-m_1) \\
+ \sum\limits_{m_1 = 0}^{M-1}\sum\limits_{m_2 = m_1}^{M-1} h_{2}(m_1,m_2)x(n-m_2)x(n-m_1)
\end{aligned}
\label{003}
\end{equation}
where the expanded input vector and filter coefficient vector can be defined by
\begin{equation}
\begin{aligned}
{\bm x}_e(n) \triangleq&\; \left[\bm x^\mathrm{T}_1(n),\bm x^\mathrm{T}_2(n)\right]^\mathrm{T}\\
=&\; [x(n),...,x(n-M+1),\\
&x^2(n),x(n)x(n-1),...,x^2(n-M+1)]^\mathrm{T},
\label{004}
\end{aligned}
\end{equation}
\begin{equation}
\begin{aligned}
{\bm h} \triangleq&\; [h_{1}(0),h_{1}(1),...,h_{1}(M-1),\\
&h_{2}(0,0),h_{2}(0,1),...,h_{2}(M-1,...,M-1)]^\mathrm{T}.
\label{005}
\end{aligned}
\end{equation}

To obtain a stable performance in a high impulsive process, for example, $\alpha=$1.1 or 1.2, two novel algorithms, termed as VFxlogLMP and VFxlogCLMP, were proposed \cite{lu2016adaptive}. These algorithms are derived from the $l_p$-norm of logarithmic cost and outperform than the conventional Volterra FxLMS (VFxLMS) and Volterra filtered-x least mean $l_p$-norm (VFxLMP) algorithms. Furthermore, it is shown in \cite{he2019an} that, by using the maximum correntropy criterion (MCC), the Volterra filtered-x recursive maximum correntropy (VFxRMC) algorithm can obtain good nonlinear capability with normalized and adaptive kernel size schemes.

\begin{figure}[!htb]
    \centering
    \subfigure[]{
    \includegraphics[scale=0.85] {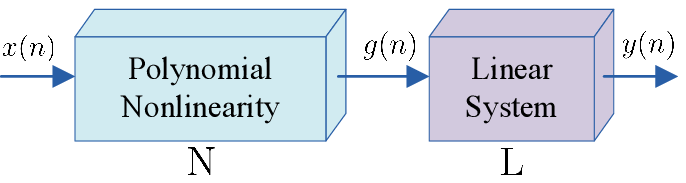}}
    \label{fig:subfig:aa}
    \subfigure[]{
    \includegraphics[scale=0.85] {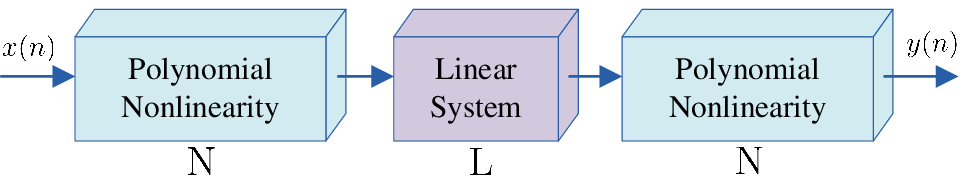}}
    \label{fig:subfig:bb}
    \subfigure[]{
    \includegraphics[scale=0.85] {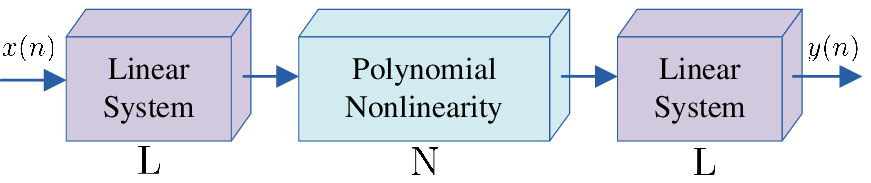}}
    \label{fig:subfig:cc}
    \caption{\label{02} Structure of the nonlinear model, where N denotes the nonlinear part and L denotes the linear system. (a) Hammerstein model; (b) Hammerstein-Wiener model; (c) Wiener-Hammerstein model.}
    \label{Fig02}
\end{figure}

\subsection{Hammerstein ANC algorithms}

As the type of point-wise nonlinear expansions, the Hammerstein system model is regarded as the class of truncated Volterra filters. The basic Hammerstein system (N-L) model is comprised of a cascade connection of a memoryless nonlinearity in series with a linear system (see Fig. \ref{Fig02} (a)). The input-output mapping of the memoryless nonlinearity can be expressed in the form of polynomial expansion
\begin{equation}
\begin{aligned}
g(n)=p_1 x(n) + p_2 x^2(n) + \ldots + p_M x^M(n)
\label{006}
\end{aligned}
\end{equation}
where $p_j$, $j=1,2,\ldots,M$ denotes the $j$th order coefficient. Combining the Wiener model (L-N), Hammerstein-Wiener model (N-L-N) is closer to the actual nonlinear system than Hammerstein model. The structure of Hammerstein-Wiener model is shown in Fig. \ref{Fig02} (b). Although Hammerstein-Wiener type nonlinear system has one more nonlinear link than Hammerstein and Wiener models, its essential feature is still a series of static nonlinear link and dynamic linear link. Changing the cascading order of the static nonlinear part and the dynamic linear part, the Wiener-Hammerstein model (L-N-L) can be obtained, as shown in Fig. \ref{Fig02} (c). The above mentioned models have been applied to NLANC system, see \cite{sra2017nl}.

\subsection{FLANN-based algorithms}

\begin{figure}[!htb]
    \centering
    \includegraphics[scale=0.8] {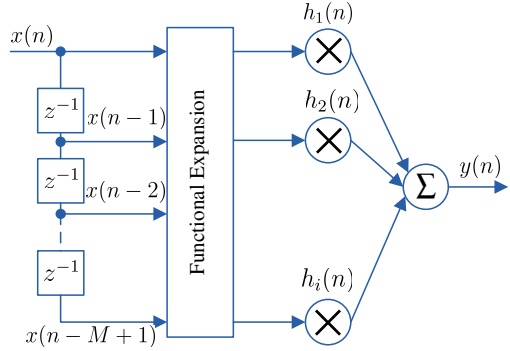}
    \caption{\label{03} Structure of the FLANN.}
    \label{Fig03}
\end{figure}

\textit{1) FsLMS-based algorithms:} Since the classical FsLMS algorithm was proposed in 2004, many FsLMS-type algorithms integrated with the FLANN controller were presented in the past decade, of which the trigonometric expansion is the most commonly used in FLANN. Fig. \ref{Fig03} shows the structure of the FLANN, where $h_i(n)$ is the coefficient of the weight vector at time instant $n$. The reference signal and weight vector can be defined as follows \cite{reddy2009fast}:
\begin{equation}
\begin{aligned}
{\bm x}_e(n) \triangleq&\; \{x(n),\mathrm{sin}[\pi x(n)],\mathrm{cos}[\pi x(n)],...,\\
&\mathrm{sin}[b\pi x(n)],\mathrm{cos}[b\pi x(n)],...,x(n-M+1),\\
&\mathrm{sin}[\pi x(n-M+1)],\mathrm{cos}[\pi x(n-M+1)],...\\
&\mathrm{sin}[b\pi x(n-M+1)],\mathrm{cos}[b\pi x(n-M+1)]\}^\mathrm{T},
\label{007}
\end{aligned}
\end{equation}
\begin{equation}
{\bm h}(n) \triangleq [h_1(n),h_2(n),...,h_{M(2b+1)}(n)]^\mathrm{T}
\label{008}
\end{equation}
where $b$ is the order of the FLANN filter. The number of coefficients of the FLANN filter can be calculated by $N_c = M(2b+1)$. From this result, we can see that the main advantage of FLANN-based algorithms is that it owns less computational burden than the Volterra ANC algorithm.

In \cite{reddy2009fast}, the FsLMS algorithm was devised for multi-channel NLANC. To achieve moderate computational complexity, the data block processing is usually preferred. In \cite{kumar2010block}, a fast Fourier transform (FFT)-based block FsLMS algorithm was developed. A RFsLMS algorithm was proposed in \cite{george2012a}, which minimizes a new logarithmic cost to obtain enhanced performance in both Gaussian and impulsive noise environments. Moreover, it extended the estimation in \cite{bergamasco2012active} to online update parameter. To further enhance performance, two novel NLANC algorithms were proposed based on the $q$-gradient method \cite{yin2018functional}. One of the algorithms was designed for Gaussian noise with time-varying $q$ scheme; the other algorithm was intentionally designed for AINC.

\textit{2) Generalized FLANN:} Most of the works focus on developing new FLANN algorithms, and little attention is paid for improving FLANN itself. In \cite{sicuranza2011a}, an interesting trial was conducted by adding suitable cross-terms between the input signal $x(n)$ and ${\rm {sin}}(b\pi x(n))$ ${\rm {cos}}(b\pi x(n))$ with different time shifts, resulting in generalized FLANN (GFLANN). Following this main idea, other improved GFLANNs were also developed in \cite{luo2018improved,le2019hierarchical}.

\textit{3) Exponential FLANN:} Inspired by Taylor series expansion that contains all odd powers of sine signal and includes all even powers for cosine signal, an exponential FLANN (EFLANN) filter was proposed for NLANC, which adds the exponential factor in trigonometric functional expansion to accelerate convergence speed \cite{patel2016design}. After that, some algorithms were developed for further enhancement of the EFLANN, such as generalized EFLANN \cite{le2018generalized} and approximated EFLANN \cite{deb2019design}.

\begin{figure}[h]
    \centering
    \includegraphics[scale=0.8] {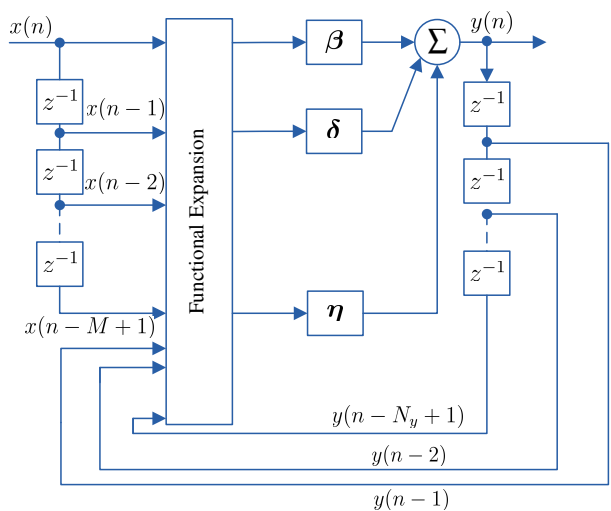}
    \caption{\label{04} Structure of the RFLANN, where $\bm \beta$, $\bm \delta$, and $\bm \eta$ are the parameter vector and $N_y$ is the recent samples of the output signal.}
    \label{Fig04}
\end{figure}

\textit{4) Recursive FLANN:} Recalling linear ANC systems, the filtered-u LMS (FuLMS) algorithm is capable of updating coefficients of the infinite impulse response (IIR) filter. In \cite{george2012development}, the IIR filter with FuLMS algorithm was convexly combined with an FsLMS filter, resulting in the FsuLMS algorithm and allowing performance enhancement with respect to its component filters. Based on a bounded-input bounded-output (BIBO) stability condition, the related work \cite{sicuranza2011bibo} has confirmed that the FuLMS algorithm can also apply for NLANC. In this case, the recursive FLANN (RFLANN) filter was proposed that uses past output as part of the input signal. The block diagram of the RFLANN is shown in Fig. \ref{Fig04}, where $\bm h^\mathrm T=[\bm \beta^\mathrm T, \bm \delta^\mathrm T,\ldots,\bm \eta^\mathrm T]$.  As a flexible extension of LIP filters, the RFLANN has become another important method. An application of the even mirror Fourier nonlinear (EMFN) filter to RFLANN has been considered for NLANC \cite{sicuranza2013new}.

\subsection{Orthogonal basis function-based algorithms}

The above Volterra ANC algorithms are derived by Volterra series  that do not satisfy
orthogonal properties. In a comparison, the orthogonal property of these LIP filters can ensure fast convergence of gradient descent algorithms, thereby improving performance over non-orthogonal LIP filters. In what follows, we review the advances of some orthogonal basis function-based algorithms since 2009.

\textit{1) Chebyshev ANC algorithms:} The Chebyshev nonlinear (CN) filter based on Chebyshev polynomials, which hold two-fold characteristics as follows. I) Chebyshev polynomials are a family of orthogonal polynomials. II) Chebyshev polynomials converge faster than expansions in other sets of polynomials. The Chebyshev polynomials are given by the generating function \cite{chen2019nonlinear}
\begin{equation}
T_{q+1}(x) = 2xT_q(x) - T_{q-1}(x),
\label{009}
\end{equation}
where $x \in [-1,1]$, $T_0(x)=1$, and $q$ is the order of Chebyshev polynomials expansion. The closed-form expression for Chebyshev polynomials of any order is expressed as
\begin{equation}
T_q(x) = \sum\limits_{i = 0}^{[\frac{q}{2}]} {(-1)^i} \binom{q}{2i}x^{q-2i}(1-x^2)^i
\label{010}
\end{equation}
where $[q/2]$ denotes the integer part of $(q/2)$. The corresponding function expansion using Chebyshev polynomials (increased from 1 to $Q$) is expressed as (\ref{011}).

The novel CN filter was proposed in \cite{chen2019nonlinear}, whose input of CN filter is derived from  correlated empirical mode decomposition (CEMD) method. As shown in \cite{chen2019nonlinear}, such Chebyshev ANC algorithm outperforms the known FsLMS and VFxLMS algorithm in noise reduction and convergence rate.

\textit{2) Fourier ANC algorithms:} Despite the FLANN and GFLANN filter significantly reduce the computational complexity, they do not satisfy the condition of the Stone-Weierstrass theorem. Therefore, they cannot be universal approximators of Volterra filters, i.e., these types of filters are unable to model arbitrary systems. In \cite{carini2012new}, a Fourier nonlinear (FN) filter was investigated for NLANC, where trigonometric basis functions are designed to avoid repetition or cancellation terms. However, as discussed in \cite{carini2014fourier}, the discontinuities on the borders of the unit hypercube cause relevant oscillations in the Fourier expansion of continuous function.

To tackle this challenge, the EMFN filter was proposed for approximation, by formulating the truncation of a multidimensional even symmetric generalized Fourier series \cite{carini2014fourier}. For the second-order EMFN filter, the input signal $\bm x_e(n)$ is expanded to $N_c = \frac{M(M+3)}{2}$ coefficients as (\ref{012})\footnote{The length of SOV filter has the same calculation as the second-order EMFN filter.}.
\begin{figure*}[!t]
\begin{equation}
\begin{aligned}
\bm x_e(n) =&\;\left\{1,T_1[x(n)],T_1[x(n-1)],\ldots,T_1[x(n-M+1)], \right.T_1[x(n)]T_1[x(n-1)],\\
&\ldots ,T_1[x(n-M+2)]T_1[x(n-M+1)], T_2[x(n)],T_2[x(n-1)],\ldots,T_2[x(n-M+1)],\\
&\ldots \left. ,T_Q[x(n)],T_Q[x(n-1)],\ldots,T_Q[x(n-M+1)]\right\}^{\mathrm T}
\end{aligned}
\label{011}
\end{equation}
\begin{equation}
\begin{aligned}
\bm x_e(n) = &\left\{ \sin\left[\frac{\pi}{2}x(n)\right],\sin\left[\frac{\pi}{2}x(n-1)\right],\ldots, \right.
\sin\left[\frac{\pi}{2}x(n-M+1)\right],\cos\left[\pi x(n)\right],\\
&\ldots,\cos\left[\pi x(n-M+1)\right],\sin\left[\frac{\pi}{2}x(n)\right]\sin\left[\frac{\pi}{2}x(n-1)\right],\\
&\ldots,\sin\left[\frac{\pi}{2}x(n-M+1)\right]\sin\left[\frac{\pi}{2}x(n-M+1)\right],\ldots,\\
&\sin\left[\frac{\pi}{2}x(n-M+3)\right]\sin\left[\frac{\pi}{2}x(n-M+1)\right],\ldots,
\left. \sin\left[\frac{\pi}{2}x(n)\right]\sin\left[\frac{\pi}{2}x(n-M+1)\right]\right\}^{\mathrm T}
\end{aligned}
\label{012}
\end{equation}
\begin{equation}
\begin{aligned}
\bm x_e(n) =&\;\{L_0[x(n)],L_1[x(n)],\ldots,L_Q[x(n)],\ldots,L_1[x(n-M+1)],L_2[x(n-M+1)],\\
&\ldots,L_Q[x(n-M+1)]\}^{\mathrm T}
\end{aligned}
\label{013}
\end{equation}
\end{figure*}
The EMFN filter was successfully employed to solve NLANC problems. The work \cite{guo2019improved} embedded the EMFN filter inside of the RFLANN structure under the BIBO stability condition. Numerical simulation results indicated that above modified EMFN filters are feasible and effective for solving NLANC problems. The work in \cite{guo2018sparse} developed two novel EMFN and CN filters with sparse modeling of nonlinear secondary path, which can significantly reduce computational load without sacrificing the control performance. Moreover, it showed that the EMFN filter outperforms the CN filter with sparse modeling.

\textit{3) Legendre ANC algorithms:} The Legendre nonlinear (LN) filter is also an orthogonal universal approximators, which shares closed properties with FN and EMFN filters. The Legendre polynomial can be obtained by the following formula \cite{george2013active}:
\begin{equation}
L_{q+1}(x) = \frac{1}{q+1}\left[(2q+1)xL_{q}(x) - qL_{q-1}(x)\right]\;\;q>2,
\nonumber
\end{equation}
where $L_q(x)$ denotes the $q$th order Legendre polynomial derived from the solution of the differential equation $\frac{\rm d}{{\rm d}x}\left[(1-x^2)\frac{{\rm d}y}{{\rm d}x}\right]+q(q+1)y=0$ and $L_0(x)=1$. The expanded input vector of the LN filter is shown in (\ref{013}). In 2013, a new algorithm was suggested in \cite{george2013active}, which uses a cascade structure of FLANN and LN filters. Since the LN filter requires $x \in [-1,1]$ for stability, a function ${\mathrm {tanh}(\cdot)}$ is used to link the output of FLANN and the input of the LN filter. Simulation results demonstrated improved performance as compared to the FsLMS algorithm \cite{george2013active}.

\begin{figure}[!htb]
    \centering
    \includegraphics[scale=0.4] {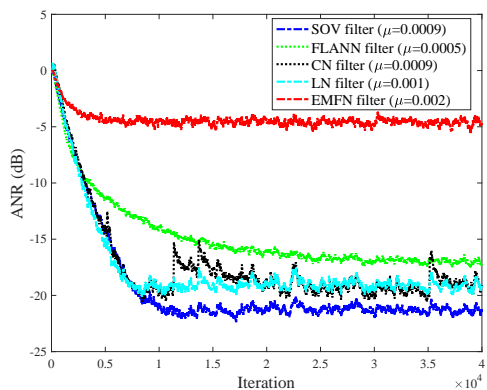}
    \caption{Averaged noise reductions (ANRs) of the representative NLANC algorithms, where the filter length of all algorithms is set to 77 and simulation settings are the same as Example 1 in \cite{yin2018functional}.}
    \label{Simu}
\end{figure}

We compare representative NLANC algorithms in Fig. \ref{Simu} for a scenario described in \cite{yin2018functional}, where the LMS algorithm is utilized to update weight vector, and $\alpha=2$ is used as the reference signal. The results show that the SOV filter has the best performance, followed by the LN filter and the CN filter, which have comparable noise residual performance.

\subsection{Bilinear ANC algorithms}

\begin{figure}[!htb]
    \centering
    \includegraphics[scale=0.8] {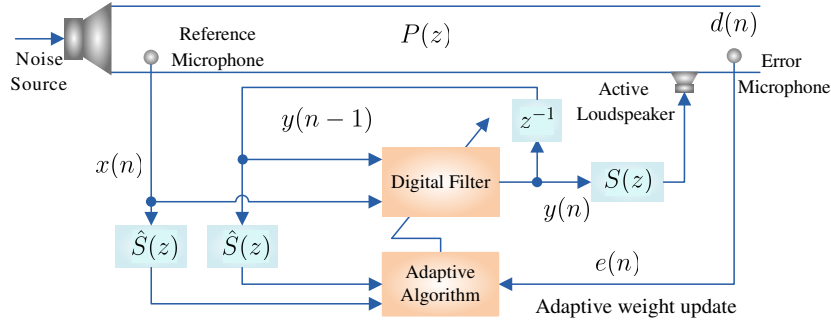}
    \caption{\label{05} Structure of the bilinear filter.}
    \label{Fig05}
\end{figure}

The deteriorated performance of the SOV filter is achieved when strong nonlinear distortions occurred in the ANC system \cite{le2019m-max}. The bilinear filter is developed to address this problem with shorter filter length. The input-output relationship of the bilinear filter is given by
\begin{equation}
\begin{aligned}
y(n) =&\; \sum\limits_{i=0}^{N_1} a_i(n)x(n-i) + \sum\limits_{t=1}^{N_1} b_t(n)y(n-t) \\
&\sum\limits_{i=0}^{N_1}\sum\limits_{t=1}^{N_1}c_{i,t}(n)x(n-i)y(n-t)
\end{aligned}
\label{014}
\end{equation}
where $a_i(n)$, $b_t(n)$, and $c_{i,t}(n)$ are the coefficients with $N_1$, $N_1+1$ and $N_1(N_1+1)$ elements, respectively. The block diagram of the convention bilinear filter for ANC problem is shown in Fig. \ref{Fig05}.

As can be seen, the bilinear filter has $x(n)$, $y(n)$, and $x(n-i)y(n-t)$ with different time shifts. An update method, namely diagonal-channel structure, was proposed in \cite{tan2014nonlinear}. This structure updates the weights of these terms independently to acquire improved performance. By using this structure, the FxLMS and filtered-x recursive least square (FxRLS) algorithms were proposed and a counterpart with sequential channel update was developed \cite{tan2014nonlinear}. Using the properties of the diagonal-channel can also reduce the computational complexity of the bilinear filter \cite{guo2020bibo,dong2020diagonal}. In \cite{tan2016implementation}, two diagonal strategies were devised for the bilinear filter in the context of NLANC system, which omits some diagonal elements to significantly reduce the computational complexity by sacrificing performance of the algorithm.

In recent years, the FLANN-inspired bilinear filters \cite{guo2018adaptive,zhu2019reweighted,le2018bilinear,le2019m-max,luo2018novel} received much attention owing to their good modeling capability. For the ANC system with saturation nonlinearity, a general function expansion bilinear filter with a 3-D diagonal-channel structure was proposed, where a trigonometric-based FLANN expansion diagonal-channel structure bilinear filter is presented \cite{guo2018adaptive}. Aimed at improving the performance of a control system with a nonlinear secondary path, a FxLMS algorithm using reweighted bilinear filters was proposed and a generalized filter to model different nonlinear secondary paths was explored in \cite{zhu2019reweighted}. In \cite{le2018bilinear}, the FLANN was introduced to the bilinear filter for NLANC, which adapts the coefficients by diagonal-channel structure and achieves refined performance than the FLANN and GFLANN. Then, the leaky bilinear filter with filtered-error LMS (FeLMS) and PU scheme was subsequently developed to enhance stability and reduce computational complexity \cite{le2019m-max}. The number of coefficients for the NLANC algorithms based on nonlinear expansion is outlined in Table \ref{Tablebb2}, where $L_F$ denotes the FIR memory size and $\mathcal N_c$ denotes the number of coefficients of the order $Q$.

\begin{table}[]
    \scriptsize
    \centering \doublerulesep=0.05pt
    \caption{Number of coefficients for the algorithms.}
    \begin{tabular}{l|l|l}
        \hline
        \textbf{Nonlinear model}    & \textbf{Number of coefficients} & \textbf{Special instruction} \\ \hline
        Volterra filter    &$N_c = \frac{(M+Q)!}{M!Q!}-1$                       & SOV filter: $N_c = \frac{M(M+3)}{2}$           \\ \hline
        Hammerstein filter &$N_c = M+L_F$                          &\begin{tabular}[c]{@{}l@{}}The nonlinear part of Hammerstein filter does\\ not require historical input and output information \end{tabular}                   \\ \hline
        FLANN filter             &$N_c = M(2b+1)$                      &Based on trigonometric expansions            \\ \hline
        GFLANN filter          &$N_c = 2M+1$                          &Generally, $b=1$ in the GFLANN filter                  \\ \hline
        EFLANN filter            &$N_c = M(2b+1)+1$                           &Require exponential operation                     \\ \hline
        CN filter   &$N_c = 1+Q(M+1)+M$                          &Generally, $Q \leq 3$ in the CN filter                      \\ \hline
        FN filter     &\begin{tabular}[c]{@{}l@{}}$\mathcal N_c = 2\sum\limits_{j=0}^{Q-1}\tbinom{Q-1}{j}\tbinom{M+j}{Q}$\\(for $Q$th order)\end{tabular}
                                  &\begin{tabular}[c]{@{}l@{}}Second-order ENFN filter: $N_c = \frac{M(M+3)}{2}$ \\  ENFN filter is suitable for modeling strong nonlinearities  \end{tabular}             \\ \hline
        LN filter    &$N_c = QM+1$                       &\begin{tabular}[c]{@{}l@{}}Suitable for modeling mild and medium\\ nonlinearities\end{tabular}            \\ \hline
        Biliner filter     &$N_c = N_1^2+3N_1+1$          &\begin{tabular}[c]{@{}l@{}}Suitable for modeling strong
        nonlinearities\\ with saturated signals \end{tabular}                     \\ \hline
    \end{tabular}
\label{Tablebb2}
\end{table}

\subsection{Discussion}

In this subsection, we discuss some existing strategies in NLANC systems. Most of these schemes have been reviewed in Part I of this work, and can be extended to solve the ANC problem under nonlinear distortions.

The remote microphone technique (RMT) has been used for NLANC in the past decade. In \cite{spiriti2014gradient}, the RMT is incorporated into the empirical weight update (EWU) algorithm to circumvent the slow convergence problem of stochastic approximation methods and PSO algorithms. Based on FLANN controller, a novel FeLMS algorithm with RMT was proposed in \cite{das2012adjoint}, which exhibits its computational advantages and reliable performance for the sinusoidal signal. Its FsLMS counterpart also has been proposed in \cite{das2012nonlinear}.

As stated in Part I, the convex combination scheme can substantially improve the performance of linear ANC. Such a strategy is also applicable to NLANC systems. The first attempt of convex combination for NLANC can be found in \cite{george2014convex}. In this work, two different nonlinear modeling methods, FLANN and Volterra, were performed by convex combination to obtain enhanced performance.

The MCC not only has been used to the Volterra FxRLS algorithm \cite{he2019an}, but also has been applied to normalized versions of the FsLMS algorithm \cite{kurian2017robust}. As a generalization of the minimum error entropy (MEE) criterion, R{\' e}nyi's entropy has been applied to NLANC. A generalized filtered-x integrated-evaluation-criterion-based gradient descent (GFx-IECGD) algorithm was presented, whose update equation is mixed by the mean square error (MSE) criterion and R{\' e}nyi's entropy \cite{zhang2010adaptive}.

The chaotic noise commonly exists in NLANC system, and some researchers developed various algorithms to control such type of noise. The work in \cite{behera2014functional} considers ANC of a mixture of tonal and chaotic noise via the hybrid FLANN method. The effort in \cite{behera2014nonlinear} first explored the characteristic of logistic chaotic noise and applied FLANN for active chaotic noise control. It was observed that the logistic chaotic noise 1 is highly broadband, while the logistic chaotic noises 2$\sim$6 are mostly broadband with some frequency notches \cite{behera2014nonlinear}. To enhance readability, we summarize these schemes in Table \ref{Table01}.

\begin{table}[htp]
    \scriptsize
    \centering \doublerulesep=0.05pt
    \caption{Contributions of the algorithms in Section \ref{sec:2}.}
    \begin{tabular}{c|c|c}
        \hline\hline
        \textbf{RMT}                    & \begin{tabular}[c]{@{}l@{}}\textbf{Convex Combination}\\ \textbf{algorithms}\end{tabular} & \begin{tabular}[c]{@{}l@{}}\textbf{Entropy-based}\\ \textbf{algorithms}\end{tabular} \\ \hline \cite{spiriti2014gradient,das2012adjoint,das2012nonlinear}
        &\cite{george2012development,george2014convex,guo2019convex}                 &\cite{he2019an,kurian2017robust,zhang2010adaptive}                                                                    \\ \hline
        \begin{tabular}[c]{@{}l@{}}\textbf{Chaotic signal}\\ \textbf{control}\end{tabular}
        & \textbf{Leaky algorithms}
        & \begin{tabular}[c]{@{}l@{}}\textbf{PU-based} \\ \textbf{algorithms}\end{tabular}     \\ \hline
        \begin{tabular}[c]{@{}l@{}}\cite{guo2019convex,le2019m-max,le2018bilinear,luo2018novel}\\ \cite{das2012nonlinear,behera2014functional,behera2014nonlinear,luo2017novel}\end{tabular}
        &\cite{le2019m-max,luo2018novel}                    &\cite{le2019m-max}                                                                    \\ \hline\hline
    \end{tabular}
\label{Table01}
\end{table}

\section{Heuristic-based ANC algorithms}
\label{sec:3}
Both linear and nonlinear ANC tasks can be solved by transforming into a global optimization problem. Especially for NLANC systems, such problems are well-suitable for considering global optimization problems, because the nonlinear secondary path may cause the conventional adaptive filter to suffer from the local minima trapping phenomenon in non-convex or non-deterministic polynomial (NP)-hard problems, and it turns out the degraded reduction performance. The first heuristic algorithm used to optimize ANC was the genetic algorithm (GA) in 1994 \cite{wangler1994genetic}. Afterward, the PSO algorithm was applied to the NLANC problem in 2006. The relevant time line can be referred to Table \ref{Table000}.

\textit{1) Genetic algorithms:} The GAs are inspired by Darwin's evolution theory, which seeks the optimal solution by studying the evolutionary process of simulation. In \cite{chang2010active}, the adaptive GA (AGA)-based ANC system was designed for various environments. Besides, this work confirmed that the AGA can suppress the nonlinear noise interference without estimating the secondary path. The use of the interior-point method (IPM) for GA-ANC has also been proved to be an effective technique \cite{raja2018bio}. The IPM searches for the optimal solution by traversing the internal feasible region, and it shows excellent performance for solving linear or quadratic constraint optimization problems. Benefiting from this, GA-ANC systems can further enhance reduction performance for various noises.

\textit{2) Backtracking search algorithms:} Closer to the GA is the backtracking search algorithms (BSA), which are a population-based iterative evolutionary algorithm (EA) designed to prevent the algorithm from falling into a local optimum. In \cite{khan2018backtracking}, the BSA integrated with sequential quadratic programming (SQP) was proposed for ANC systems with sinusoidal signal and complex random signal, which can obtain improved performance without modeling $S(z)$.

\textit{3) PSO algorithms:} Based on the competition and cooperation mechanism between the particle swarm, the PSO can produce a swarm intelligence to guide the optimization search. By exploiting the soft bound ${\rm {tanh}}\{\cdot\}$ as the saturation nonlinearity module of the secondary path, the PSO has been applied to ANC in \cite{rout2016particle}. Specifically, it can tackle the cases with varying saturation grades of both reference and secondary paths. Similar to the GA, the PSO-ANC can also reduce noise without exact estimation of the secondary path, see \cite{rout2012particle,rout2019pso,raja2019design}. As a nonlinear FLANN extension of \cite{rout2012particle}, the algorithm in \cite{george2012particle} was trained by PSO for multi-channel NLANC systems, which is subsequently developed into a decentralized version. In the PSO-literature, a less common exploration strategy is the Wilcoxon norm \cite{george2012robust2}, which has been proved to have effectiveness for data contaminated by outliers. It has been shown through simulation experiments in \cite{george2012robust2} that considerable reduction can be obtained as compared with the MSE criterion.

\textit{4) Bacterial foraging optimization algorithms:} The bacterial foraging optimization (BFO) is based on the foraging behavior of \textit{escherichia coli} bacteria in human intestine. Bacteria look for nutrients to maximize the energy gained in a unit of time. They go through different stages of chemotaxis, swarming, reproduction and elimination-and-dispersal. Such BFO algorithms have been successfully employed to adapt the coefficients of the NLANC controller, which outperforms the standard GA-ANC by nearly 5dB in steady-state \cite{gholami2012non,gholami2014active1}.

In addition to the above mentioned heuristic algorithms, other heuristic algorithms have been studied in the ANC system. The firefly (FF) algorithm was a classical optimization and has been used to NLANC. The new algorithm cascades the FLANN and FIR filter, and adopts the FF algorithm to obtain coefficients \cite{walia2018design}. The fireworks algorithm (FWA) is also a swarm intelligence algorithm inspired by the fireworks explosion. The research in \cite{khan2019a} reveals the profound optimization capacity of the three variants of FWA for NLANC.

\section{Novel NLANC methods emerging in past decade}
\label{sec:4}
In this section, we present some novel NLANC approaches, such as spline ANC algorithms, KAF ANC algorithms, and nonlinear distributed ANC algorithms, proposed in the last ten years.

\subsection{Spline ANC algorithms}
\begin{figure}[!htb]
    \centering
    \includegraphics[scale=0.8] {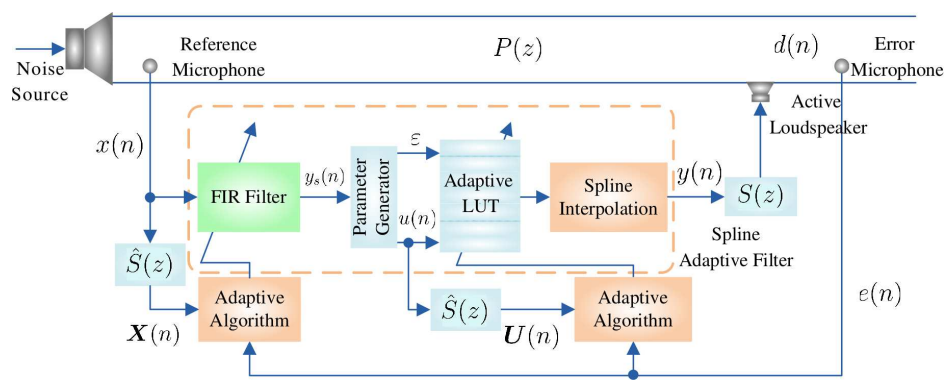}
    \caption{\label{06} Structure of the NLANC system with spline adaptive filter.}
    \label{Fig06}
\end{figure}

In 2013, the spline adaptive filter was originally proposed for nonlinear system identification, whose adaptation is based on a spline function \cite{scarpiniti2013nonlinear}. As shown in \cite{scarpiniti2013nonlinear} and references therein, the spline adaptive filter enjoys good nonlinear modeling performance in diverse applications and owns low computational cost. Such spline adaptive filter was extended to address NLANC problem in \cite{patel2015nonlinear}. The schematic of the NLANC system with a spline adaptive filter is shown in Fig. \ref{Fig06}, where $\bm U(n)$ is the filtered signal of $\bm u(n)=[u^3(n),u^2(n),u(n),1]^{\mathrm T}$, and $\varepsilon$ denotes the span index, and $u(n)$ is the local parameter, which can be computed by
\begin{subequations}
    \begin{equation}
    \varepsilon = \left\lfloor \frac{y_s(n)}{\Delta x} \right\rfloor + \frac{C-1}{2}
    \end{equation}
    \begin{equation}
    u(n) = \frac{y_s(n)}{\Delta x} - \left\lfloor \frac{y_s(n)}{\Delta x} \right\rfloor
    \label{015ab}
    \end{equation}
\end{subequations}
where $y_s(n)$ denotes the output of the finite impulse response (FIR) filter, $\Delta x$ denotes the gap between the control points, $C$ denotes the total number of control points in the activation function, and $\lfloor \cdot \rfloor$ stands for the floor operation. Then, the generated parameters pass through the adaptive look-up table (LUT) and are interpolated by a local low order polynomial spline curve. Hence, the output of the spline adaptive filter is
\begin{equation}
y(n) = \bm u^{\mathrm T}(n) \bm \Omega \bm q
\label{016}
\end{equation}
where $\bm \Omega$ is the pre-computed spline basis matrix and $\bm q$ is the control point vector.

As an initial step towards multi-channel spline adaptive filters, the algorithm proposed in \cite{patel2020multi}  has obvious advantages in terms of computational complexity and final MSE level when compared to the multi-channel VFxLMS and FsLMS algorithms. Furthermore, the advantages of the FIR spline adaptive filter can also refer to \cite{patel2016design2}. In 2016, a novel infinite impulse response (IIR) spline adaptive filter was proposed \cite{patel2016compensating}, which corresponds to the use of a feedback scheme and the FuLMS algorithm. Compared with the FIR filters, the main merit of using IIR filters is their implementation efficiency.

\subsection{KAF-ANC algorithms}

\begin{figure}[!htb]
    \centering
    \includegraphics[scale=0.8] {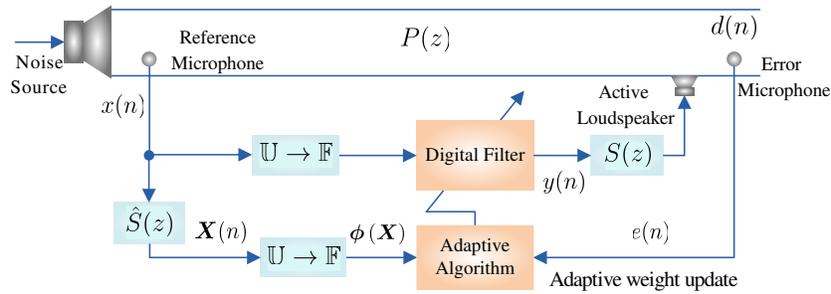}
    \caption{\label{07} Structure of the NLANC system with KAF.}
    \label{Fig07}
\end{figure}
The kernel trick has become rather popular and has been successfully applied to adaptive filtering. The key idea of the KAF is to recast the input data from input space $\mathbb U$ into a high-dimensional feature space $\mathbb F$ via a reproducing kernel Hilbert space (RKHS). As a result, the inner product operations in linear adaptive filters are translated to the calculation of a kernel function $\kappa\left(\cdot\right)$ in the feature space without knowing the exact nonlinear mapping.

The KAF was first applied to NLANC system in 2009 \cite{bao2009active}. The basic block diagram of the ANC with KAF is shown in Fig. \ref{Fig07}, where $\bm \phi: \mathbb U \to \mathbb F$ denotes feature vector. By using \textit{Mercer theorem}, the inner products in RKHS can be calculated as
\begin{equation}
\kappa\left(\bm {X},\bm {X'}\right) = \bm \phi\left(\bm {X}\right)\bm \phi^{\mathrm T}\left(\bm {X'}\right)
\label{017}
\end{equation}
where ${\bm {X}},{\bm {X'}} \in \mathbb U$ and the kernel function $\kappa\left(\cdot\right)$ often uses Gaussian kernel, which is expressed as $\kappa\left(\bm {X},\bm {X'}\right) = {\exp}\left(-\varsigma\left\|{\bm {X}}-{\bm {X'}}\right\|^2\right)$, where $\varsigma$ is the kernel size and $\left\|\cdot\right\|$ denotes the $l_2$-norm. Finally, the filter output can be expressed as
\begin{equation}
y(n) = \sum\limits_{j = 1}^n {{a_j} \langle \bm \phi (\bm X),} \bm \phi (\bm X_j)\rangle
\label{019}
\end{equation}
where $a_j$ denotes the coefficient and $\langle\cdot\rangle$ denotes the inner product operation. The work in \cite{bao2009active} considered the scenario that chaotic noise and only the primary path $P(z)$ is nonlinear, which has certain limitations. The reference \cite{liu2018kernel} made up for this shortcoming, which took the nonlinear secondary path into account. Moreover, some multi-tonal and sinusoidal noise sources are simulated. The choice of the kernel function is crucial to the performance of the kernel algorithm, and most works employed Gaussian kernel as a default choice due to its universal approximating capability, numerical stability, and desirable smoothness. The work in \cite{raja2019novel} specifically discussed the selection of different kernel functions. Three novel kernel functions, namely logistic kernel, tan-sigmoid kernel, and inverse-tan kernel, were proposed and subsequently applied to KAF for NLANC. Its simulation results demonstrated the improved performance by using new kernel functions.

It should be noted that the above mentioned KAF and spline adaptive filters have been exhibited improved performance as compared to the conventional FsLMS and VFxLMS algorithms \cite{patel2015nonlinear,liu2018kernel}. Hence, we can conclude that the KAF and spline adaptive filters are good candidates for NLANC.

\subsection{Nonlinear distributed ANC algorithms}
\begin{figure}[!htb]
    \centering
    \includegraphics[scale=0.7] {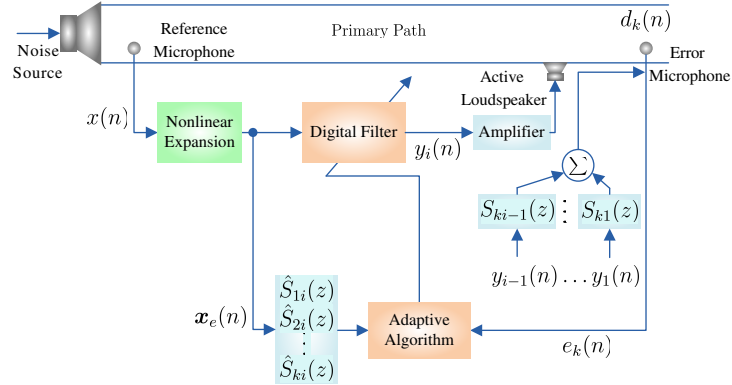}
    \caption{\label{08} Structure of the nonlinear distributed multi-channel ANC system.}
    \label{Fig08}
\end{figure}

The idea of a nonlinear distributed ANC system is basically derived from a linear distributed ANC network. The difference lies in the use of function expansion strategies. In \cite{kukde2017on}, a nonlinear distributed multi-channel ANC algorithm was proposed followed by the diffusion strategy, where the LN filter is employed and both primary and secondary paths are considered to be nonlinear. Fig. \ref{Fig08} shows the structure of the nonlinear distributed multi-channel ANC algorithm, where $i=1,2,\dots,I$, $I$ denotes the number of loudspeaker, $k=1,2,\dots,K$, and $K$ denotes the number of error sensors. Very recently, a new nonlinear distributed ANC system was presented by using incremental implementation and adaptive EFLANN expansion, which exhibits lower computational cost than the distributed ANC system with the FsLMS, VFxLMS, and conventional EFLANN controller \cite{kukde2020incremental}.

\section{Recent implementations and applications of ANC}
\label{sec:5}

    \begin{table}[]
        \scriptsize
        \centering \doublerulesep=0.05pt
        \caption{Time line of ANC applications.}
            \begin{tabular}{l|l|l|l}
                \hline\hline
                \textbf{Years} & \textbf{Authors} & \textbf{Contributions} & \textbf{References} \\ \hline
                1993      &Craig and Angevine          &\begin{tabular}[c]{@{}l@{}}Active transformer noise \\control\end{tabular}                &\begin{tabular}[c]{@{}l@{}} \cite{craig1993active}\end{tabular}            \\ \hline
                1994      &Sutton et al.          &\begin{tabular}[c]{@{}l@{}}ANC for road noise\end{tabular}                &\begin{tabular}[c]{@{}l@{}} \cite{sutton1994active}\end{tabular}            \\ \hline
                1997      &McJury et al.          &\begin{tabular}[c]{@{}l@{}}ANC for fMRI\end{tabular}                &\begin{tabular}[c]{@{}l@{}} \cite{mcjury1997use}\end{tabular}            \\ \hline
                2002      &Gan and Kuo         &\begin{tabular}[c]{@{}l@{}}Integrated feedback ANC\\for headset\end{tabular}                &\begin{tabular}[c]{@{}l@{}} \cite{gan2002integrated}\end{tabular}            \\ \hline
                2002      &Qiu et al.          &\begin{tabular}[c]{@{}l@{}}Waveform synthesis method\\for active transformer noise\\ control\end{tabular}                &\begin{tabular}[c]{@{}l@{}} \cite{qiu2002waveform}\end{tabular}            \\ \hline
                2003      &Jakob and M{\"o}ser          &\begin{tabular}[c]{@{}l@{}}ANC with double-glazed \\windows\end{tabular}                &\begin{tabular}[c]{@{}l@{}} \cite{jakob2003active,jakob2003active2}\end{tabular}            \\ \hline
                2003      &\begin{tabular}[c]{@{}l@{}}Pinte, Desmet,\\ and Sas\end{tabular}           &\begin{tabular}[c]{@{}l@{}}Iterative learning control \\algorithm for ANC systems\end{tabular}                &\begin{tabular}[c]{@{}l@{}} \cite{pinte2003active}\end{tabular}            \\ \hline
                2005      &Zhou et al.          &\begin{tabular}[c]{@{}l@{}}Back-propagation (BP)\\-FeLMS algorithm\end{tabular}                &\begin{tabular}[c]{@{}l@{}} \cite{zhou2005analysis}\end{tabular}            \\ \hline
                2006      &\begin{tabular}[c]{@{}l@{}}Kuo, Mitra,\\ and Gan\end{tabular}          &\begin{tabular}[c]{@{}l@{}}Feedback ANC for headphone\end{tabular}                &\begin{tabular}[c]{@{}l@{}} \cite{kuo2006active}\end{tabular}            \\ \hline
        \hline\hline
        \end{tabular}
        \label{Table001}
    \end{table}

In this section, the application development of ANC in the past decade is discussed. Before we start, it should be highlighted that many applications have been developed before last decade (see time line of ANC applications in Table \ref{Table001}). These works have laid the foundation for the implementations and applications of ANC in the past ten years.

\subsection{ANC implementation}

\textit{1) DSP implementation:} ANC techniques have been implemented in real-time systems such as digital signal processor (DSP), field programmable gate array (FPGA), and very large scale integration (VLSI). Several hardware implementations dedicated to studying lower-power consumption and small area realization have been reported. The compensated-FxLMS algorithm implemented by fixed-point DSP was presented in \cite{chang2009efficient}. However, such method requires a specialized measurement of the devices in the control system. In addition, the round-off and quantization errors will change with the programmer's code implementation, and the correct values of the phase and amplitude compensation factors need to be determined \cite{kim2017enhancement}. An improvement of fixed-point DSP method was proposed in \cite{kim2017enhancement}, which added feedback and \textit{communication error}\footnote{The definition of communication error can be found in Eq. (12), Part I of this work.} Considering the implementation of different structures in fixed-point DSP, a detailed performance comparison of FxLMS, FxNLMS, IIR, and subband ANC algorithms was presented in \cite{siravara2003comparative}. The research results showed that the FxLMS and its variant are more suitable for ANC implementation and the fixed-point DSP has about 2dB degradation as compared with its floating-point counterpart.

\textit{2) FPGA implementation:} The DSP owns limited processing capability and it is intended for low sampling rate applications. In contrast, the performance of the FPGA implementation largely depends on the selection of FPGA device. Several algorithms and architectures were proposed for performance enhancement with high-speed computational architecture \cite{ardekani2011filtered,shi2017multiple,leva2011fpga}. The novel algorithm, called filtered-weight FxLMS (FwFxLMS), was proposed \cite{ardekani2011filtered}. The FwFxLMS algorithm can efficiently increase the upper-bound of the step size and as such, accelerating the convergence rate of algorithm without affecting the stability. Devoted to reducing computational load of multi-channel ANC system, a novel architecture of FPGA was proposed to solve the conflicted requirement between throughput and hardware resource consumption by using parallel execution and folding \cite{shi2017multiple}.

\textit{3) VLSI implementation:} The VLSI provides a large amount of calculation in a small area, and the resource assumptions is significantly reduced compared to FPGA and DSP. The effort in \cite{mohanty2017hardware} focused on VLSI implementation on the FxLMS and FsLMS algorithms. Prior to this, few literature considered the implementation of NLANC algorithms. By devising folded structures of the algorithm, the intentionally proposed delay FsLMS algorithm becomes easy to use in practice.

\subsection{ANC through open window}

ANC through open windows technique can attenuate noise while maintaining natural ventilation in dwellings \cite{kwon2013interior,huang2011active,wang2019boundary,wang2019boundary2,murao2019hybrid}. The finite-element-based ANC methods have been used to establish the control performance and the passive attenuation offered by the open window \cite{lam2018active,lam2018physical}. In \cite{lam2018active}, an ANC system with 16-channel and secondary sources distributed evenly across the opening was implemented with a full-sized window, yielding more than 5dB below 2000Hz overall attenuation. However, the size of an opening influences the control performance and the opening windows may also introduce noise at the same time \cite{elliott2018wavenumber}. To avoid this issue, a scheme of small opening in the wall was implemented in \cite{zhang2020secondary}. To design an ANC system with reduced computational complexity of DSP implementation and acceptable level of noise reduction, a novel multi-channel ANC method was proposed for opening window. By preprocessing the error signals, the number of inputs to the controller can be decreased and a good performance can be achieved when it employs 4-channel ANC system with 0.2$\times$0.2m opening window \cite{murao2017mixed}.

\subsection{Headphone application}

Many headphone applications of ANC were developed based on DSP, FPGA and VLSI implementations \cite{vu2015low}. Some algorithms focus on implementing ANC headphones with less resources, head-tracking problems, multi-channel, and multi-reference \cite{jung2017combining,behera2017head,cheer2019application}. The most popular algorithm used in ANC headphone is the feedback FxLMS algorithm. A number of (in-ear) headphones based on the feedback ANC system were implemented \cite{chang2010active2,zhang2013intuitive}. Their performance comparison of feedback ANC system with different circuit implementations can be seen in \cite{vu2017high}. In \cite{schumacher2011active}, a hybrid feedback ANC scheme was presented by making use of the combination of classical and adaptive feedback ANC techniques. Considering that system delay will limit implementation performance in practice, this work also proposed a mixed analog-digital realization to maintain very low system delay in the classic part and enable adaptive ANC components to adapt to the digital domain.

In comparison, a feedforward structure was adopted for binaural ANC system in \cite{tanaka2017binaural}. It is worth pointing out that the performance improvement of \cite{tanaka2017binaural} also benefits from the use of the parametric array loudspeakers (PAL) as the secondary sources. Similar techniques have been applied to ANC problems such as steerable PAL \cite{tanaka2010active} and curved-type PAL \cite{tanaka2011mathematically}. Also considering the use of a binaural ANC system, a combined bilateral-binaural ANC (CBBANC) system was proposed for closed-back headphones \cite{belyi2020combined}. By switching between the binaural and bilateral implementation, the improved performance can be obtained as compared to the conventional bilateral ANC headphones.

The ANC headphone application has also embedded some techniques discussed above. For example, in \cite{das2013nonlinear}, the virtual sensing method was adopted. Particularly, in \cite{das2013nonlinear}, the FsLMS algorithm with nonlinear virtual microphone was developed for active headrest application, which considers the nonlinear distortion in the presence of primary noise. The prediction filter was another method applied in feedback ANC headphones, which does not need any real-time coefficient updates and is therefore very economical \cite{guldenschuh2013prediction}.

\subsection{Zone-of-quiet based approach}

The traditional ANC technique can only minimize the noise level at the location where the error microphone is placed. Correspondingly, a zone-of-quiet (ZoQ) is created around this error microphone. However, unless multiple sensors are used, it is not technically possible to control or monitor noise levels in ZoQ. The applications of spherical loudspeakers arrays for ANC has been presented to extend ZoQ \cite{rafaely2009spherical}. Following the idea of \cite{rafaely2009spherical,peleg2011investigation}, several improvements were proposed including linear array \cite{hart2012active}, and RMT-based methods \cite{elliott2015modeling}. To design larger ZoQ, a method was proposed by introducing one and two secondary sources, which aimed at maximizing the area of the 10dB quiet zones \cite{tseng2011local}. By introducing feedforward adaptive ANC structure to control shape and extension of ZoQ, a new avenue was proposed in \cite{ardekani2014adaptive}, which has capability of creating a controllable spherical quiet zone via a compensation factor. Moreover, in \cite{wrona2018shaping}, a ZoQ method equipped with memetic algorithm (MA) and distributed adaptation has been proposed, which holds high potential for practical applications.

The ZoQ based methods have been integrated into ANC for speech enhancement in hearing aids. In \cite{serizel2009zone}, the ANC with noise reduction strategy was proposed for a specific ZoQ. Following a different direction, the method in \cite{serizel2012zone} developed two schemes. The former method is based on the MSE criterion at a remote point, and the latter method is based on an average MSE criterion over a desired ZoQ. Moreover, several approaches on integrated ANC and noise reduction for hearing aids have been proposed, such as those proposed in \cite{serizel2009integrated,pradhan2017speech,albu2018efficient} and references therein.

\subsection{Spatial ANC approach}

Spatial ANC is designed to attenuate noise within a specific space. Since there is not only one space point but the space must be controlled, many transducers are needed. A direct method to achieve spatial ANC is to employ the multi-channel FxLMS algorithm. However, such method cannot guarantee that the spatial noise is also reduced, and the calculation cost is high. To address this drawback, several mode-domain methods were proposed in \cite{zhangg2016sparse,zhang2018active,zhang2018spatial,maeno2018mode}. Such methods can significantly reduce computational complexity since the mode-domain conversions can reduce the cross-correlations between transducers. In addition, the filter update can be computed by a mode-independent manner. Very recently, a new spatial ANC method was developed, where the higher-order source (HOS) scheme is used to improve controllability of efficiency and error \cite{murata2019global}.

\subsection{fMRI acoustic noise control}

The problem of fMRI acoustic noise control is a research focus in the application of ANC technology. Studies showed that when magnetic resonance imaging (MRI) scans, it can generate acoustic noise up to 130dB \cite{reddy2011hybrid}. So far, a large number of ANC algorithms were proposed for fMRI problem, see \cite{kumamoto2011active,kannan2010efficient} and references therein. The followed articles implemented by head-mounted ANC system with piezoelectric loudspeakers and optical microphones \cite{kida2009head} or RMT \cite{miyazaki2015head}. The effectiveness of the FxRLS-FxNLMS algorithms on fMRI noise control was validated that the hybrid algorithm can gain a very fast convergence \cite{reddy2011hybrid}. This group then proposed an improved parallel feedback ANC system \cite{ganguly2015improved}, which parallels two FxNLMS algorithms with the NLMS decomposition to achieve exceeding 40dB noise attenuation level.

The adaptive comb filters (ACFs) that perform a coordinated search, have been applied to fMRI noise control. Based on the previous works, i.e., SONIC (self-optimizing narrowband interference canceller) \cite{niedzwiecki2009new,niedzwiecki2009new2,niedzwiecki2009self}, the ACF incorporated SONIC algorithm was suggested for active fMRI noise control \cite{niedzwiecki2013estimation}. This algorithm can track the fundamental frequency and the amplitude of different frequency components of non-stationary harmonic signals embedded in white measurement noise. However, its reduction level is limited to 20dB. For performance improvement, a discrete Fourier transformation (DFT)-based frequency-domain ANC algorithm for fMRI noise control was proposed in \cite{lee2017frequency}, which can achieve approximately 35dB overall reduction in the frequency range of interest.

\subsection{Periodic noise control}

Based on different characteristic distributions with time, the exhaust noise can be divided into three types: steady, periodic, and intermittent \cite{zhou2013optimal}. Among the three types, the periodic noise is the most widely encountered and has great research significance. The repetitive impulsive noise belongs to a category of periodic noise and has many practical backgrounds. For example, when compressed air escapes regularly through the exhaust pipe, it generates repetitive impulsive noise, up to 120dB. In the past decade, many studies have focused on and analyzed this kind of noise. Similarly, the FxLMS algorithm and its variants can also be used to attenuate this noise \cite{sun2015convergence}.

The iterative learning control (ILC) algorithm is another method that can effectively control repetitive impulsive noise, which can be regarded as the adaptive feedforward algorithm. In ILC controllers, the control signal is adapted to the specific repetitive of transient noise. It runs repetitively in an open-loop over a finite time interval to reduce the same discrete interference at each time instant. Related to the ILC method is the well-known repetitive control (RC) algorithm, which can also be interpreted as the adaptive feedforward algorithm. The ILC and RC algorithms share the same core idea, that is, it should be possible to improve the performance of a system that repeatedly performs the same repetitive task. The difference is that the RC algorithm has continuously active property.

It demonstrated in \cite{pinte2010novel,stallaert2010novel} that the drawbacks of conventional ILC and RC methods are high computational complexity and low robustness of the controllers. To solve these problems, a novel strategy to design non-causal ILC and RC filters was described, with emphasis on the application of ANC. It can achieve 4dB noise reduction with the frequency-domain ILC filter. Owing to the fact that the ILC and RC algorithms are closely related, many design methods can be used in both algorithms. For example, in \cite{zhou2013active}, a non-causal transversal FIR filter was combined with the optimal ILC algorithm for active repetitive impulsive noise control, and then it was applied to the optimal RC algorithm \cite{zhou2013optimal}.

\subsection{Transformer noise control}
Transformer noise is typically characterized by even and multiples of the line frequency of the fundamental and harmonic frequencies. Generally, the fundamental frequency does not vary by more than $\pm$0.5\% \cite{zhang2012active}. Therefore, using the internally synthesized preset frequency signal as a reference signal for transformer noise control may be a reasonable way. In \cite{zhang2012active}, the classical FxLMS algorithm with an internally synthesized reference signal was proposed, which can obtain a maximum of 15dB noise reduction. Motivated by the online and offline secondary path modeling strategy, a novel FxLMS algorithm was presented for 50000KVA transformer noise control \cite{zhao2017new}. Simulations demonstrated that an 84.10$\sim$96.86\% decrease in average sound energy density can be obtained in a certain area.

\subsection{ANC in vehicles}
ANC in vehicles requires a good spatial and temporal matching between the fields of the actuator or secondary sources \cite{samarasinghe2016recent}. Spatial matching requires that the wavelength of the acoustic wave for active noise control must be small, whereas time matching demands the system can adapt to changes in vehicle speed and load \cite{elliott2010active}. Owing to the fact that ANC has a significant improvement, particularly at low-frequency and on vehicles with specific problems, such technique has been widely used in vehicles \cite{cheer2012active}. In the past ten years, many prominent companies and corporations have proposed effective solutions to this problem, see references in \cite{samarasinghe2016recent}. In \cite{cheer2014design}, three feedback ANC systems were investigated based on internal model control (IMC) architecture and frequency discretization method. Moreover, the multi-channel  algorithm has addressed ANC of road noise in vehicles \cite{cheer2015multichannel,jung2019local}. In \cite{jung2019local}, eight reference sensors located around the four wheels were devised to obtain a broadband reduction of 4dB up to 1kHz. In \cite{guo2018active}, a variable step-size (VSS) median-least mean square algorithm was presented for active rail vehicle interior noise control. In \cite{cheer2016active} and \cite{peretti2013adaptive}, the leaky FxLMS and multi-channel FxLMS algorithms have been respectively applied to ANC for yacht.

The ANC technique is quite general and its application is not limited to the above mentioned works. The ANC principle has already been used for mobile phones \cite{cheer2018application}, earmuff systems \cite{moon2015performance,zhu2019design}, high-speed elevators \cite{yang201423}, mufflers \cite{chen2009simulations}, and washing machines \cite{mazur2019active}. In \cite{rohlfing2014ventilation,gardonio2014modular}, the single and multi-channel ANC systems have been employed for ventilation duct. Furthermore, new problems in the ANC system, such as broadband noise fields in rooms \cite{ma2019active}, electroencephalogram (EEG) signal analysis \cite{bagha2018understanding} have also been studied. These applications show the great application potential of ANC.

We note that a number of papers on NLANC assume the presence of nonlinearities that are difficult to encounter in real-world applications. Therefore, most of above mentioned works are based on linear assumptions and do not consider nonlinear characteristics. To clearly show the benefits offered by a nonlinear approach with respect to the linear one, the implementations and applications of NLANC are summarized in Table \ref{Table003}.
\begin{table}[]
    \scriptsize
    \centering \doublerulesep=0.05pt
    \caption{Implementations and applications of NLANC in Section \ref{sec:5}.}
    \begin{tabular}{l|l|l}
        \hline
        \textbf{Contributions}                & \textbf{Benefits}                                          & \textbf{References} \\ \hline
        FsLMS implementation           &\begin{tabular}[c]{@{}l@{}}The proposed delayed block FsLMS (DBFsLMS)\\ structure involves 58\% less area-delay product\\ and 41\% less energy per sample than the FxLMS\\ structure. \end{tabular}         &\cite{mohanty2017hardware}                     \\ \hline
        \begin{tabular}[c]{@{}l@{}}Generalized filter bank based\\nonlinear ANC algorithm\end{tabular}           &\begin{tabular}[c]{@{}l@{}}The new algorithm is suitable for a headrest\\ application with head movement, and the effective\\ area of active headrest is achieved as high as\\ 45 cm $\times$ 45 cm. \end{tabular}         &\cite{behera2017adaptive}                     \\ \hline
        \begin{tabular}[c]{@{}l@{}}NLANC with virtual\\ microphone control\end{tabular}
        &\begin{tabular}[c]{@{}l@{}}These methods can control noise at a remote\\ location when the system is nonlinear.\end{tabular}
        &\cite{das2012nonlinear,das2012adjoint,das2013nonlinear}                     \\ \hline
    \end{tabular}
    \label{Table003}
\end{table}

\section{Future research challenges}
\label{sec:6}

Some future challenges of ANC have been discussed in \cite{kuo2010active,krish}. In this section, we present a number of promising directions for future ANC research from other perspectives.

\subsection{Theoretical analysis of algorithms}

As we discussed in Part I of this work, the fractional lower order moment (FLOM)-based algorithms can be effectively used for AINC. The problem of interest is to analyze the theoretical performance of the FLOM-based algorithms in the presence of $\alpha$-stable noise. However, such a topic is extremely difficult since the variance of $\alpha$-stable noise is infinite, leading to some assumptions does not satisfy the conditions. As far as we know, researchers did not consider the case in which the error in $\alpha$-stable noise environment is changing across iterations. Moreover, the steady-state and transient performance of conventional least mean $p$th power (LMP) algorithm has not been investigated in the presence of $\alpha$-stable noise. The lack of these analysis foundations causes the analysis of the FLOM-based algorithms exactly difficult.

The similar problem also exists in the analysis of LIP algorithms, such as the FN and EMFN. Moreover, since the input vectors of these algorithms are expanded by different basis functions, the behavior of algorithms is more complicated.

\subsection{Sparsification for KAF-ANC}

Despite the above KAF algorithms are well applied in NLANC, there are still problems to be solved. For system identification problem, several KAF algorithms with curb of network size were proposed. Since the network size of the KAF linearly increases with the number of training data, which means that the filter length is very high during adaptation. The problem of NLANC is the same as that of KAF for system identification. This increase in complexity can be harnessed through the application of some sparsification techniques, for example, the quantized scheme in \cite{chen2011quantized} and the set-membership scheme in \cite{flores2019set-membership}. These strategies can curb the number of input data into the kernel function at each iteration. If it is detected that not enough new information is carried in the input signal, it will be abandoned as the redundant data.

\subsection{The application of Internet of Things to ANC}

IoT attracts significant attention in the fields of location, smart transportation, and signal processing. Based on the concept of IoT, the ANC technique has been applied to IoT. By utilizing wireless communication with acoustics, IoT was brought into the ANC context, which provided a potential solution to urban noise pollution \cite{shen2018mute}. Through DSP implementation verification, the proof of concept (PoC) system was stated, which allows the concept of IoT to be applied in the field of ANC \cite{galambos2015active}. In this work, the analog cables were replaced by digital ethernet transmission, resulting in a more flexible system that is easier to be rearranged and modified.

It is noted that existing ANC-IoT schemes rely on the single-node device, and the impact of practical constraints, such as impulsive noise, the nonlinear distortions, and distributed implementation, on the performance of ANC-IoT has not been investigated yet. Furthermore, as discussed before, ANC can effectively suppress the noise in vehicles. Hence, ANC may be specifically applied to the Internet of Vehicles (IoV).

\subsection{Discussion}
Currently, NLANC has made much progress. In the following, some future directions that are peculiar to NLANC are presented.

1) For NLANC algorithms, most of the algorithms are still at the theoretical and laboratory stage, and they are not fully utilized in practical applications. For the future directions of nonlinear algorithms in NLANC, it needs to curb its computational complexity. Since the nonlinear filtering problem is inherently infinite dimensional, the substantial increase of data samples can prohibit practical applications of NLANC algorithms. Therefore, low complexity NLANC algorithms are an important research direction in the future.

2) Selecting the optimal memory length of the NLANC algorithm for the corresponding nonlinear systems is also a research direction. In the field of echo cancellation, the memory control scheme of Volterra filter has been studied  \cite{zeller2010adaptive}. However, it has not been investigated in the NLANC adaptive filtering algorithms.  Such a problem is also applicable to the heuristic-based algorithms in NLANC. The optimal size of populations for heuristic-based algorithms is of great significance to the reduced computational complexity and improved performance.

3) Further exploration of NLANC technology in specific practical scenarios are required. Currently, NLANC does not seem to find its most suitable breakthrough point for large-scale commercial applications. Thus, finding a cost-effective application scenario is also an important future research direction of NLANC.

\section{Conclusion}
\label{sec:7}
In this survey paper, we have discussed state-of-the-art methods related to NLANC technique, including FsLMS-based algorithms, distributed extension and selected applications. We have also summarized FLANN approaches as well as the Chebyshev, EMFN, and LN filters. The spline adaptive filter, kernel adaptive filter, and nonlinear distributed ANC algorithms are the new nonlinear modeling approaches considered for NLANC. Due to technical limitations, these open areas need further investigations of sophisticated NLANC techniques. The review of implementations for both linear and nonlinear ANC are merged together. We have also discussed the main challenges for future ANC techniques.

\section*{Acknowledgment}
\label{sec:Ack}
The authors would like to thank the associate editor and the anonymous referees for their
valuable comments.

\section*{References}

\bibliography{mybibfile2}

%

\end{document}